\documentclass{article} 
\usepackage{tencent_tech_report}
\usepackage[colorlinks = true,
            linkcolor = blue,
            urlcolor  = blue,
            citecolor = blue,
            anchorcolor = blue]{hyperref}

\usepackage{enumitem} 
\usepackage{wrapfig}
\usepackage{authblk}
\usepackage{lipsum}
\usepackage{fontawesome5} 
\usepackage{algorithm}
\usepackage{algorithmicx}
\usepackage{algpseudocode}
\usepackage{multirow}
\usepackage{pifont}  
\usepackage[most]{tcolorbox}

\newtcolorbox{alprompt}[1]{
        boxrule = 1pt,
        fontupper = \small\tt,
        fonttitle = \bf\color{black},
        arc = 2pt,
        rounded corners,
        colframe = black,
        colbacktitle = white!97!yellow,
        colback = white!97!yellow,
        title = #1,
}

\newtcolorbox{promptbox}[2][Prompt]{%
    breakable,                
    colback=black!5!white,
    colframe=#2,
    arc=5pt, 
    boxrule=0.5pt,
    fonttitle=\bfseries,
    title=#1, 
    before upper={\small},
    fontupper=\fontfamily{ptm}\selectfont,
}

\usepackage{amsfonts}       
\usepackage{nicefrac}       
\usepackage{lipsum}
\usepackage{natbib}
\usepackage{subcaption}
\usepackage[table,xcdraw]{xcolor}

\usepackage{wrapfig}
\usepackage{graphicx}
\usepackage{amsmath,amssymb,amsfonts}
\usepackage{textcomp}
\usepackage{xcolor}
\usepackage{listings}

\usepackage{cite}
\usepackage{microtype}
\usepackage{hyperref}
\usepackage{url}
\usepackage{booktabs}
\usepackage{enumitem}
\usepackage{multicol}
\usepackage{makecell}
\usepackage{CJKutf8}
\usepackage{amsmath}
\usepackage{siunitx}
\usepackage{floatflt}
\usepackage{wrapfig}
\usepackage{graphicx}
\usepackage{booktabs}
\usepackage{wrapfig}
\usepackage{authblk}
\usepackage{lipsum}
\usepackage{fontawesome5} 

\usepackage{algorithm}
\usepackage{algorithmicx}
\usepackage{algpseudocode}
\usepackage{microtype}
\usepackage{graphicx}
\usepackage{multirow}
\usepackage{booktabs} 
\usepackage{pifont}  
\usepackage{graphicx}  
\usepackage{subcaption} 

\algnewcommand{\LeftComment}[1]{\Statex \(\triangleright\) #1}

\usepackage{array}
\usepackage{amsmath}
\usepackage{amssymb}
\usepackage{mathtools}
\usepackage{amsthm}
\usepackage{arydshln}
\usepackage[capitalize,noabbrev]{cleveref}
\usepackage{adjustbox} 
\usepackage{enumitem}
\usepackage{fontawesome5}

\theoremstyle{plain}

\theoremstyle{definition}

\theoremstyle{remark}

\usepackage{subfloat}

\usepackage[textsize=tiny]{todonotes}

\definecolor{nblue}{RGB}{65, 105, 225}    
\definecolor{norange}{RGB}{255, 140, 0}   
\definecolor{ngreen}{RGB}{34, 139, 34}    

\newcommand{\scriptdata}{\texttt{ScriptBench}}
\newcommand{\scriptagent}{\texttt{ScripterAgent}}
\newcommand{\directoragent}{\texttt{DirectorAgent}}
\newcommand{\criticagent}{\texttt{CriticAgent}}

\colmfinalcopy

\title{{\em The Script is All You Need:} An Agentic Framework for Long-Horizon Dialogue-to-Cinematic Video Generation}

\author[ ]{\bf Chenyu Mu\thanks{Equal Contribution.}~~$^{,1,2}$}
\author[ ]{\bf Xin He$^{*,1}$}
\author[ ]{\bf Qu Yang$^{*,1}$}
\author[ ]{\bf Wanshun Chen$^{1}$}
\author[ ]{\bf Jiadi Yao$^{1}$}
\author[ ]{\bf Huang Liu$^{1}$}
\author[ ]{\bf Zihao Yi$^{1}$}
\author[ ]{\bf Bo Zhao$^{1}$}
\author[ ]{\bf Xingyu Chen$^{1}$}
\author[ ]{\bf Ruotian Ma$^{1}$}
\author[ ]{\bf Fanghua Ye$^{1}$}
\author[ ]{\bf Erkun Yang$^{2}$}
\author[ ]{\bf Cheng Deng$^{2}$}
\author[ ]{\\ \bf Zhaopeng Tu\thanks{Correspondence to: Zhaopeng Tu \textless zptu@tencent.com\textgreater.}~~$^{,1}$}
\author[ ]{\bf Xiaolong Li$^{1}$}
\author[ ]{\bf Linus$^{1}$}

\affil[1]{Tencent Hunyuan Multimodal Department} 
\affil[2]{Xidian University \protect\\[4pt] 
\url{https://github.com/Tencent/digitalhuman/tree/main/ScriptAgent}}

\begin{document}

\maketitle

\begin{figure*}[htbp]
    \centering
    \includegraphics[width=\linewidth]{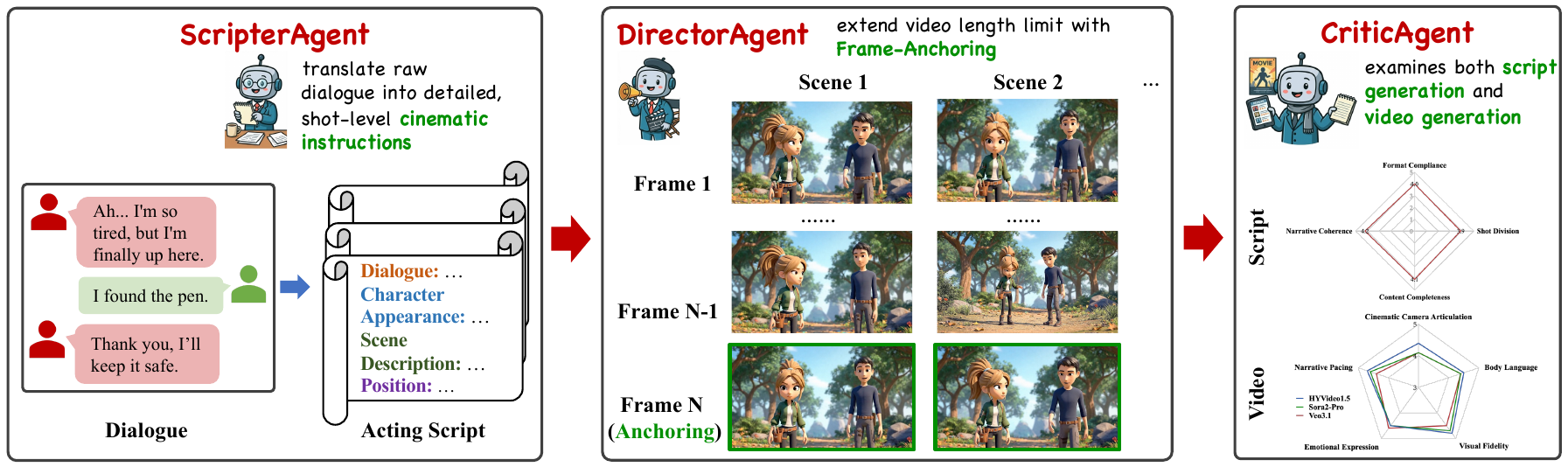}
    \caption{Our proposed pipeline consists of three key components: (1) a \scriptagent, trained with GRPO to align its outputs with professional directorial standards; (2) a \directoragent, which ensures seamless visual continuity across scenes, thereby overcoming the temporal incoherence caused by the fixed-duration constraints of SOTA video generation models; and (3) a \criticagent, which evaluates the generated film from both technical and cinematic perspectives.}
    \label{fig:pipeline}
\end{figure*}

\begin{abstract}
Recent advances in video generation have produced models capable of synthesizing stunning visual content from simple text prompts. However, these models struggle to generate long-form, coherent narratives from high-level concepts like dialogue, revealing a ``semantic gap'' between a creative idea and its cinematic execution. To bridge this gap, we introduce a novel, end-to-end agentic framework for dialogue-to-cinematic-video generation. Central to our framework is \scriptagent, a model trained to translate coarse dialogue into a fine-grained, executable cinematic script. To enable this, we construct \scriptdata, a new large-scale benchmark with rich multimodal context, annotated via an expert-guided pipeline. 

The generated script then guides \directoragent, which orchestrates state-of-the-art video models using a cross-scene continuous generation strategy to ensure long-horizon coherence. 
Our comprehensive evaluation, featuring an AI-powered \criticagent~and a new Visual-Script Alignment (VSA) metric, shows our framework significantly improves script faithfulness and temporal fidelity across all tested video models. 
Furthermore, our analysis uncovers a crucial trade-off in current SOTA models between visual spectacle and strict script adherence, providing valuable insights for the future of automated filmmaking.
\end{abstract}

\section{Introduction}

\begin{quote}
``{\em To make a great film you need three things: {\bf the script}, {\bf the script} and {\bf the script}.}''\\
{\raggedleft \small --- Alfred Hitchcock\par}
\end{quote}

The emergence of powerful video generation models like Sora2-Pro~\citep{brooks2024video}, Veo-3.1, and Wan2.5~\citep{wan2025wan} has marked a new era in artificial intelligence, demonstrating a remarkable ability to synthesize high-fidelity, realistic video clips from simple text prompts. However, this paradigm reveals a fundamental limitation when confronted with complex creative tasks. A significant yet underexplored ``semantic gap'' exists between a high-level narrative concept, such as a dialogue-driven scene, and the detailed, executable plan required to produce a cinematically coherent video. 
As the legendary director Alfred Hitchcock famously stated, ``{\em To make a great film you need three things: {\bf the script}, {\bf the script} and {\bf the script}}''. Inspired by this wisdom, we argue that the missing piece in automated filmmaking is the script itself.

This paper inverts the conventional video-language relationship from passive description (video-to-text) to active generation. We tackle a new, challenging task: given only coarse-grained dialogue, the model must \emph{anticipate} and \emph{generate} an executable filmmaking plan. This introduces three fundamental challenges: \textbf{(1) fine-grained contextual understanding} to resolve ambiguities in sparse dialogue; \textbf{(2) domain knowledge of filmmaking} to produce valid camera specifications; and \textbf{(3) creative reasoning} to bridge what is said with what must be shown. To address these challenges, we introduce a complete, agentic framework for dialogue-to-cinematic-video generation, composed of three core components: \scriptagent, \directoragent, and \criticagent.

To facilitate our research, we first construct \scriptdata, a large-scale benchmark for this task. Each instance features a rich, trimodal context (dialogue, audio, and character positions) and is annotated using a novel, expert-guided pipeline that ensures cinematic plausibility through multi-round error correction. Building on this, we develop \scriptagent, a model trained to transform dialogue into a structured cinematic script. We employ a two-stage training paradigm: supervised fine-tuning (SFT) to learn the script's structure, followed by Group Relative Policy Optimization (GRPO)~\citep{shao2024deepseekmath} with a hybrid reward function to align the model's creative choices with expert aesthetics. The resulting script is then passed to \directoragent, which orchestrates SOTA video models using a novel Cross-Scene Continuous Generation strategy with frame-anchoring to produce long-horizon, coherent videos that overcome the temporal limitations of current generators.

Our comprehensive experiments demonstrate the effectiveness of this script-centric approach. The full \scriptagent~model significantly outperforms existing methods, with human experts rating its outputs higher in both \textit{Dramatic Tension} (4.1 vs. 3.7) and \textit{Visual Imagery} (4.3 vs. 3.8). Furthermore, using our generated scripts as input universally improves the performance of all tested video models (including \texttt{Sora2-Pro} and \texttt{Veo3.1}), boosting metrics like \textit{Script Faithfulness} by up to +0.4 points. Our analysis also uncovers a critical trade-off in these models between visual spectacle and script faithfulness. To quantify temporal fidelity, we introduce a novel metric, Visual-Script Alignment (VSA), which confirms that our method enhances temporal-semantic coherence by over 7 points. This work provides the first end-to-end solution for dialogue-driven cinematic video generation and offers a new paradigm for automated storytelling.

\paragraph{Contributions.} Our contributions are summarized as follows:
\begin{enumerate}[leftmargin=12pt]
    \item We propose a novel agentic framework for the task of dialogue-to-cinematic-video generation, comprising three specialized agents---\scriptagent~(for script generation), \directoragent~(for long-horizon video execution), and \criticagent~(for evaluation)---that together bridge the semantic gap between sparse dialogue and coherent cinematic output.
    \item We introduce \scriptdata, a large-scale, high-quality benchmark with rich multimodal context, curated via a novel expert-guided pipeline. We also develop \scriptagent, a model trained with an innovative two-stage SFT and reinforcement learning paradigm to generate professional-quality cinematic scripts.
    \item We conduct a comprehensive evaluation of state-of-the-art video generation models, revealing a fundamental trade-off between visual spectacle and script faithfulness. We validate our framework's ability to improve temporal coherence with a new metric, Visual-Script Alignment (VSA), demonstrating significant gains across all tested models.
\end{enumerate}

\begin{figure*}
    \centering
    \includegraphics[width=0.92\linewidth]{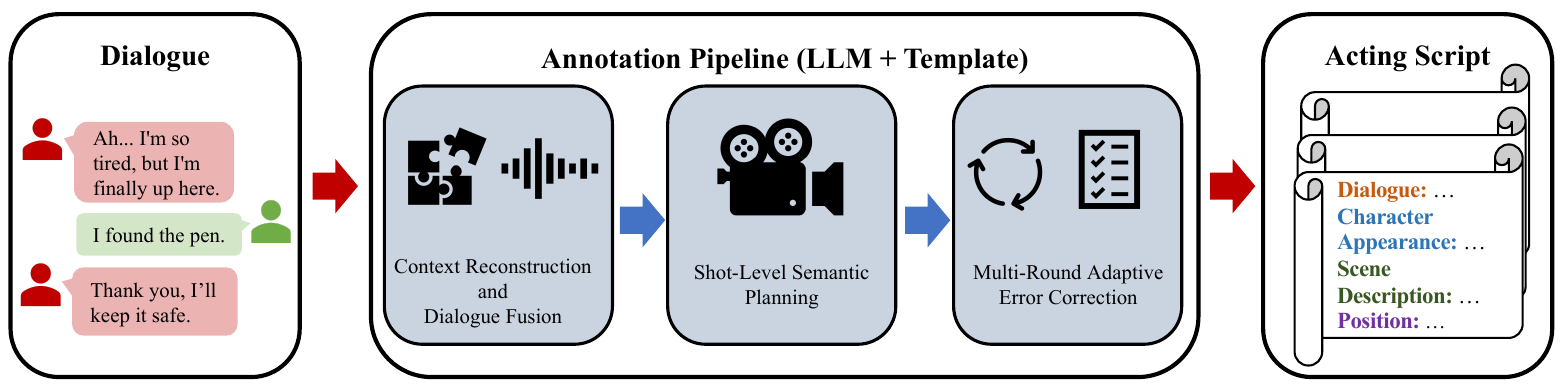}
    \caption{The three-stage, expert-guided pipeline for creating the \scriptdata.}
    \label{fig:data_preprocess}
\end{figure*}

\section{\scriptagent: Dataset and Model}

Prevailing video-language research has largely focused on the paradigm of passive description, where models learn to generate textual descriptions for existing video content (e.g., captioning or question answering). In contrast, our task inverts this relationship: given only coarse-grained dialogue, the model must \emph{anticipate} and \emph{generate} an executable filmmaking plan. This generative setting introduces three fundamental challenges absent from conventional benchmarks: 
\begin{enumerate}[leftmargin=12pt]
    \item \textbf{fine-grained contextual understanding} to resolve ambiguities inherent in sparse dialogue;
    \item \textbf{domain knowledge of filmmaking} to produce technically valid camera specifications and staging directions;
    \item \textbf{creative reasoning abilities} to bridge the gap between what is said and what should be shown.
\end{enumerate}
To bridge this gap and foster research in automated cinematic planning, we introduce \scriptdata, a new benchmark designed specifically for this task, along with \scriptagent, a dedicated model trained to transform dialogue into professional-quality cinematic scripts.

\subsection{\scriptdata}

To facilitate our study, we constructed a large-scale, high-quality dataset of cinematic scripts. 
We curated raw instances from high-fidelity cinematic cutscenes. These sources were selected for their rich dialogue, professional cinematography, and high visual consistency, which closely approximate real-world film production.

A central contribution of our work is a scalable yet high-fidelity annotation pipeline that expands sparse dialogue into rich, shot-level cinematic scripts. Although we leverage the SOTA LLM (\texttt{gemini-2.5-pro}), the entire process is tightly governed by expert-defined templates, domain constraints, and validation rules to ensure both cinematic plausibility and physical consistency. The pipeline operates in three stages:

\paragraph{Stage 1: Context Reconstruction and Dialogue Fusion.} The model first parses the multimodal inputs to reconstruct a comprehensive understanding of the scene. It jointly analyzes the textual script and dialogue audio to infer character relationships, scene settings, plot developments, emotional tendencies, and speaking intent. This process fuses the disparate signals into a coherent narrative context that makes implicit causal relations explicit.

\paragraph{Stage 2: Shot-Level Semantic Planning.} Utilizing the reconstructed context, the model plans shots under four constraints to ensure visual and narrative continuity. \textit{Shot integrity} enforces self-contained units, introducing cuts only upon clear camera or scene changes. \textit{Duration adaptation} caps shots at 10 seconds to align with generation limits. \textit{Semantic coherence} aligns boundaries with narrative transitions (e.g., emotional shifts), while \textit{Technical feasibility} prevents segmentation during complex camera motions. These principles jointly ensure the shot units are narratively meaningful and technically viable for downstream generation.

\paragraph{Stage 3: Multi-Round Adaptive Error Correction.}
In the final stage, the system executes a multi-round, adaptive error-correction loop (Figure~\ref{fig:data_preprocess}, right) to ensure both structural validity and semantic fidelity of the generated scripts. We design four verification modules: (a) \textit{Dialogue Completeness}, which ensures that all spoken content is either explicitly transcribed or marked as \texttt{[No Dialogue]}; (b) \textit{Character Appearance Consistency}, which enforces strict adherence to predefined character descriptions; (c) \textit{Scene Coherence}, which tracks environmental elements and validates narratively justified transitions; and (d) \textit{Positional and Physical Rationality}, which verifies spatial relations against plausible blocking and camera geometry. An automated detector iteratively scans the scripts, feeds corrective signals back to the generator, and repeats the loop until all constraints are satisfied. To further validate practical reliability, professional script consultants conducted a random audit on 60\% of the generated instances, revealing that while the automated pass rate reached 94\%, expert review exposed subtle semantic errors such as character teleportation, dialogue–action conflicts, and inconsistent prop states. These findings were incorporated into refined prompt constraints and verification logic, resulting in a controlled refinement process that constitutes a key novelty of our pipeline and yields cinematic scripts that are structured, internally consistent, and grounded in long-horizon narrative and physical continuity.

\paragraph{Dataset Statistics and Usage}

This pipeline yielded 1,750 finalized script instances, each in one-to-one correspondence with a raw multimodal input. The average duration of each video clip is approximately 15.4 seconds, providing sufficient temporal scope for multi-shot sequences while remaining tractable for current generative models.
The dataset is partitioned into a training set (1700 instances) and a test set (50 instances).
This partitioning scheme intentionally challenges the model to infer complete cinematic elements from conversational content alone, emulating the real-world process where directors visualize a story from a dialogue-driven script.

\subsection{\scriptagent}

Building upon \scriptdata, we develop \scriptagent, a generative model designed to automatically transform coarse-grained dialogue into a fine-grained, structured cinematic script. While large-scale foundation models demonstrate strong general capabilities, we hypothesize that this specialized task that requires domain knowledge of shot composition, pacing, and visual continuity, can benefit from targeted training on curated data. To this end, we employ a two-stage training paradigm: supervised fine-tuning (SFT) to learn the script format and narrative structure, followed by reinforcement learning (RL) to align the model's outputs with professional directorial aesthetics.

\subsubsection{Stage One: SFT for Structural Competence}
The initial stage focuses on teaching the model the fundamental syntax and structure of cinematic scripts. We formulate this as a sequence-to-sequence task, where the input $x$ is a multi-turn dialogue from our dataset, and the output $y$ is the target script in a structured JSON format. We fine-tune \textbf{Qwen-Omni-7B} as our base model, $\pi_{\text{base}}$, chosen for its strong capabilities in long-context processing and instruction following. The training objective is to maximize the conditional log-likelihood of the ground-truth script:
\begin{equation}
\mathcal{L}_{\text{SFT}} = -\mathbb{E}_{(x,y)\sim\mathcal{D}}\left[\sum_{t=1}^{|y|} \log \pi_{\theta}(y_t | y_{<t}, x)\right]
\end{equation}
where $\theta$ denotes the model parameters. We train for 20 epochs using the AdamW optimizer with a learning rate of $\eta = 1 \times 10^{-5}$, a batch size of 4, and a maximum sequence length of 8,192 tokens. This SFT stage equips the model to generate scripts that are structurally correct and content-complete, forming a solid foundation for the subsequent creative refinement.

\subsubsection{Stage Two: RL for Cinematic Alignment}
While SFT ensures structural validity, it is insufficient for capturing the subjective artistry of professional filmmaking. As suggested in our evaluation (Table~\ref{tab:script_eval}), effective scriptwriting transcends logical correctness, involving aesthetic judgments about shot composition, pacing, and emotional impact. To bridge this gap, we introduce a reinforcement learning stage to align \scriptagent~ with expert directorial preferences. We employ \textbf{Group Relative Policy Optimization (GRPO)}~\citep{shao2024deepseekmath,zhan2026mathsmith,zhan20263viewsense}, an advanced preference alignment method whose group-based relative scoring is well-suited for creative tasks that have a subjective, one-to-many nature of valid outputs.

\paragraph{Hybrid Reward Function.} A key novelty of our RL stage is a hybrid reward function, $R_{\text{total}}$, that balances objective correctness with subjective quality. It is a weighted sum of two complementary signals, with $\alpha=0.4$:
\begin{equation}
R_{\text{total}}(y) = \alpha \cdot R_{\text{structure}}(y) + (1-\alpha) \cdot R_{\text{human}}(y)
\end{equation}
\begin{itemize}[leftmargin=12pt]
   \item \textbf{Rule-Based Structural Reward ($R_{\text{structure}}$):} This component provides an objective signal for technical correctness, mirroring the verification modules from our data annotation pipeline. It evaluates and aggregates normalized scores from four automated checks: \textit{Format Compliance} (correct JSON structure), \textit{Dialogue Completeness} (all spoken lines accounted for), \textit{Scene and Character Consistency}, and \textit{Physical Rationality} (plausible character positions and camera geometry).

   \item \textbf{Human Preference Reward ($R_{\text{human}}$):} To capture cinematic aesthetics, we model expert human judgment. A team of three senior art directors scored SFT model outputs on a 1--5 scale across four creative dimensions: shot division rationality, character acting and emotion, visual aesthetics, and directorial intent. Using ~500 such annotated samples ($\mathcal{D}_{\text{pref}}$), we trained a BERT-based regression model to predict a normalized preference score in [0, 1], serving as a scalable proxy for expert cinematic taste.
\end{itemize}

\paragraph{GRPO Optimization.} During optimization, for each input $x$, we generate $K=8$ candidate scripts $\{y^{(k)}\}_{k=1}^K$ from the current policy $\pi_{\theta}(\cdot | x)$ and calculate their rewards $R_k = R_{\text{total}}(y^{(k)})$. These rewards are then used to compute a normalized advantage within the group:
\begin{equation}
A_k = \frac{R_k - \bar{R}}{\sigma_R + \epsilon}, \quad \text{where} \quad \bar{R} = \frac{1}{K}\sum_{k=1}^K R_k
\end{equation}
The policy is updated by maximizing the advantage-weighted log-likelihood, constrained by a KL-divergence penalty to prevent large deviations from the SFT initialization:
\begin{equation}
\mathcal{L}_{\text{GRPO}} = \mathbb{E}_{x\sim\mathcal{D}}\left[\frac{1}{K}\sum_{k=1}^K A_k \cdot \log \pi_{\theta}(y^{(k)} | x)\right] - \beta \cdot \mathbb{E}_{x\sim\mathcal{D}}\left[\text{KL}\left(\pi_{\theta}(\cdot|x) \,\|\, \pi_{\text{SFT}}(\cdot|x)\right)\right]
\end{equation}
where the KL coefficient $\beta=0.04$. The model is trained for 5,000 steps using the Adam optimizer with a learning rate of $\eta=10^{-6}$ and a batch size of 4. This two-stage paradigm successfully elevates \scriptagent's capabilities from generating structurally correct scripts to producing cinematically compelling plans aligned with professional standards.

\section{\directoragent: Long-Horizon Script-to-Video Execution}
\label{sec:directoragent}

While \scriptagent\ provides a structured, shot-by-shot blueprint, translating this plan into a continuous video presents its own formidable challenge. The \texttt{DirectorAgent} is designed to bridge this execution gap, acting as an automated orchestrator that transforms the generated script into a coherent, high-fidelity video sequence. Its primary function is to overcome a fundamental limitation of current video generation models: restricted temporal capacity. State-of-the-art models are typically limited to generating short clips (e.g., 8–12 seconds), far short of the 1–3 minute duration of a complete narrative scene. Naively segmenting a script and generating clips independently leads to severe artifacts, such as identity drift, inconsistent styling, and a loss of narrative continuity.

The core novelty of the \texttt{DirectorAgent} is a \textbf{Cross-Scene Continuous Generation Strategy}, which ensures both semantic coherence and visual consistency across multiple generated segments. This strategy combines (1) intelligent, shot-aware segmentation that respects cinematographic boundaries with (2) a frame-anchoring mechanism that conditions each new segment on the final state of the previous one.

\begin{figure*}[t!]
    \centering
    \includegraphics[width=0.95\linewidth]{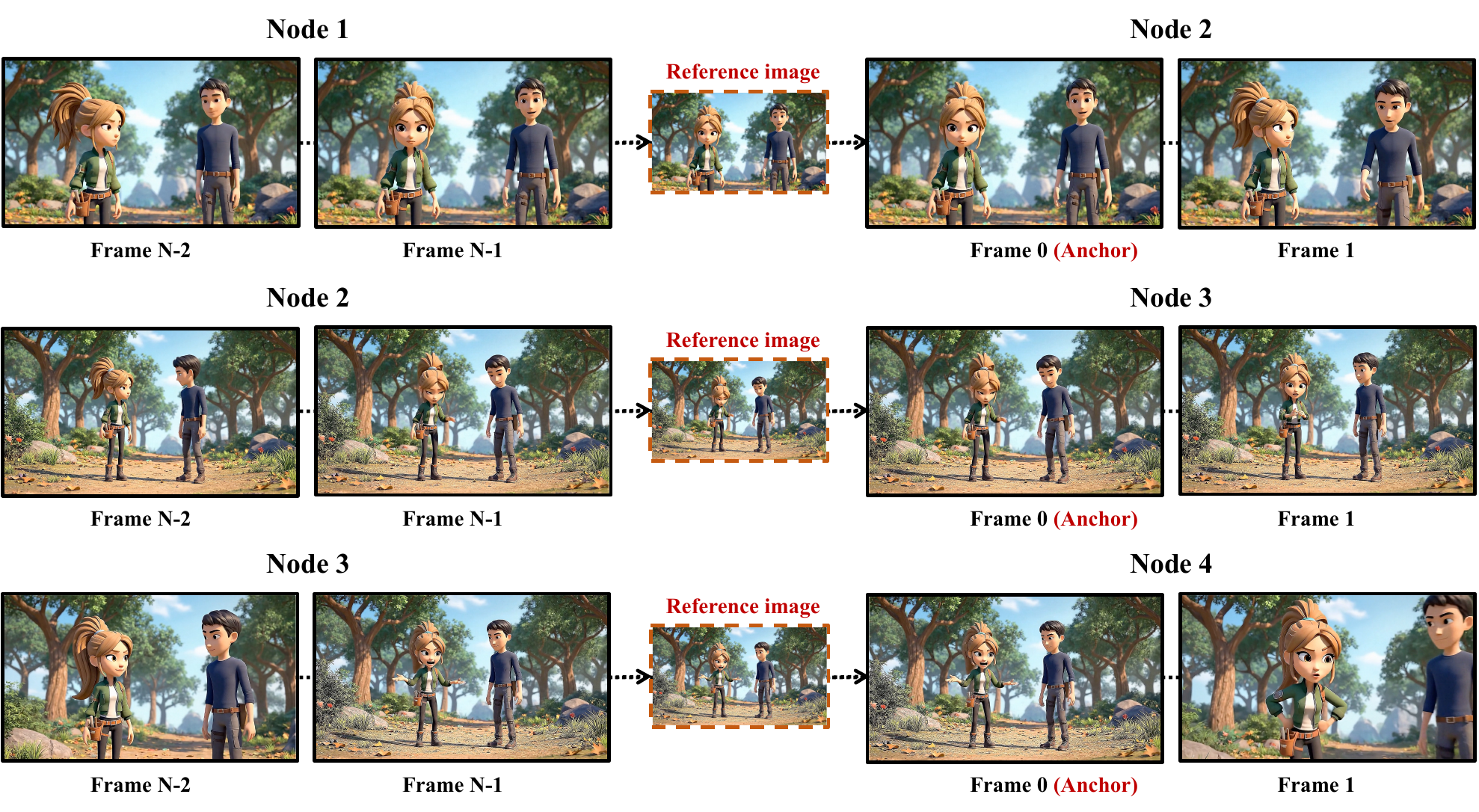}
    \caption{\textbf{Illustration of the Frame-Anchoring Mechanism.} The \texttt{DirectorAgent} utilizes the last frame of the preceding scene ($N-1$) to visually condition the generation of the current scene's first frame (0). This frame-anchoring strategy preserves visual consistency in character appearance, attire, and scene layout across multiple generation cycles, creating a seamless visual relay.}
    \label{fig:video_node_vis}
\end{figure*}

\paragraph{Intelligent Shot-Based Segmentation.}
Instead of making arbitrary temporal cuts, the agent partitions the full script into a sequence of generation tasks, or ``scenes'' that align with the natural cinematographic boundaries defined by \scriptagent. This process adheres to four key principles:
\begin{enumerate}[leftmargin=12pt]
    \item \textit{Shot Integrity:} Each scene must contain one or more complete shot units, preventing cuts in the middle of a continuous camera take.
    \item \textit{Duration Adaptation:} The total duration of a scene is constrained to fit within the target model's generation window, with a 10\% safety buffer.
    \item \textit{Semantic Coherence:} Divisions are prioritized at natural narrative breakpoints, such as the end of a character's line or a shift in emotional tone.
    \item \textit{Technical Feasibility:} Segmentation is favored at fixed camera positions, avoiding cuts during complex camera movements which are harder to transition between seamlessly.
\end{enumerate}

\paragraph{Frame-Anchored Continuity.}
To ensure seamless visual transitions between scenes, the \texttt{DirectorAgent} employs a \textbf{First-Last Frame Connection Mechanism}. As illustrated in Figure~\ref{fig:video_node_vis}, the final frame of a generated scene $i$ is extracted and used as a visual anchor or conditioning image for the generation of the subsequent scene $i+1$. This technique provides a strong visual prior for the video model, explicitly instructing it to maintain consistency in character identity, clothing, facial details, and spatial layout. By forming a visual relay from one segment to the next, this mechanism substantially reduces the identity drift and jarring scene changes that plague naive segmentation approaches. To further enhance transition quality, we also inject explicit text like ``Continuing from the previous scene'' into subsequent prompts.

\paragraph{Effectiveness and Limitations.} This strategy effectively extends the coherence window of any underlying video model. By conditioning each new segment on the visual state of its predecessor, it transforms a long-horizon generation problem into a sequence of locally solvable, continuity-preserving subproblems. While this approach significantly reduces identity drift and layout inconsistencies, challenges such as imperfect lip synchronization and residual misalignment of fine-grained actions remain.

\section{\criticagent: The Evaluation Framework}
\label{sec:evaluation}

To assess our system comprehensively, we introduce a multifaceted evaluation framework that examines both stages of our pipeline: script generation (dialogue-to-script) and video generation (script-to-video). This framework is essential because cinematic quality is inherently multidimensional, encompassing technical correctness, narrative fidelity, and subjective artistic merit. The framework combines objective metrics, automated scoring via our AI-powered \criticagent, and qualitative evaluations by human experts. All scores are assigned on a 0--5 scale. Detailed metric definitions and evaluation prompts are provided in Appendix~\ref{appendix:metrics}.

\subsection{Script Generation Evaluation}

For the dialogue-to-script stage, we evaluate the generated scripts on both their structural correctness, using our \criticagent, and their artistic quality, via a panel of human experts.

\paragraph{AI Rating (\criticagent)} We employ our \criticagent, powered by \texttt{gemini-2.5-pro}, to automatically assess generated scripts based on four criteria:
\begin{enumerate}[leftmargin=12pt]
    \item \textbf{Format Compliance.} Assesses strict adherence to the required JSON format, ensuring all key fields (e.g., ``Shot Type'', ``Camera Movement'', ``Description'') are present and correctly structured.
    \item \textbf{Shot Division Rationality.} Evaluates the logical segmentation of the script into shots, ensuring that breaks align with narrative beats and emotional shifts without being overly fragmented or lengthy.
    \item \textbf{Content Completeness.} Measures whether the script provides rich, actionable details for filming and enriches the narrative with visual information absent from the source dialogue.
    \item \textbf{Narrative Coherence.} Determines whether the sequence of shots is logically connected and if the visual storytelling flows smoothly to complement the dialogue's context.
\end{enumerate}

\paragraph{Human Rating (Directors' Panel)} To complement the automated assessment, a panel of professional directors and screenwriters evaluates the artistic quality of the scripts. They provide ratings on a 0--5 scale for three key creative aspects, which collectively indicate the script's potential for being filmed successfully:
\begin{enumerate}[leftmargin=12pt]
    \item \textbf{Character Portrayal Consistency.} Assesses whether each character's personality, speaking style, and behavior remain coherent and believable throughout the script.
    \item \textbf{Dramatic Tension \& Rhythm.} Measures the script's effectiveness in building, sustaining, and releasing dramatic tension, as well as the naturalness and engagement of its pacing.
    \item \textbf{Visual Imagery \& Cinematic Expressiveness.} Assesses how vividly the script conveys visual information and how effectively it employs cinematic language (e.g., shots, staging, atmosphere) to support the narrative.
\end{enumerate}

\subsection{Video Generation Evaluation}

We evaluate the script-to-video generation stage on two primary axes: script-video alignment and overall video quality. In addition, we conduct automatic evaluation that combines standard video quality metrics with a novel measure of script alignment.

\paragraph{AI Rating (\criticagent)} 
For the video generation stage, \criticagent evaluates the cinematic quality and faithfulness of the generated video to the source script and reference audio across five dimensions:
\begin{enumerate}[leftmargin=12pt]
    \item \textbf{Cinematic Camera Articulation.} Measures the sophistication of camera work, including shot types, framing transitions, and dynamic movements that support the scripted narrative.
    \item \textbf{Kinetic Body Language \& Blocking.} Assesses whether character motions, physical interactions, and spatial arrangements are specific, expressive, and consistent with the scripted actions.
    \item \textbf{Visual Descriptive Fidelity.} Evaluates how well the visual details (e.g., character appearance, clothing textures, scene layout, lighting) match the descriptive cues in the script.
    \item \textbf{Emotional Arc \& Micro-Expressions.} Examines whether the facial expressions, subtle gestures, and temporal evolution reflect the intended emotional progression in the script and audio delivery.
    \item \textbf{Narrative Pacing \& Timing.} Measures the alignment of shot timing, action beats, and pauses with the narrative structure and rhythm implied by the script and audio.
\end{enumerate}

\paragraph{Human Rating}
Human annotators also assess the final generated videos, providing ratings on five dimensions that collectively offer a comprehensive view of video quality:
\begin{enumerate}[leftmargin=12pt]
    \item \textbf{Visual Appeal.} Evaluates the realism, aesthetic quality, and rendering stability of the video.
    \item \textbf{Script Faithfulness.} Assesses how accurately the video adheres to the provided script in terms of scenes, actions, and plot progression.
    \item \textbf{Narrative Coherence.} Measures whether the video forms a logically consistent and easy-to-follow story, with reasonable scene transitions and pacing.
    \item \textbf{Character Consistency.} Evaluates whether characters maintain a stable identity and appearance throughout the video.
    \item \textbf{Physical Law Adherence.} Assesses whether motions and interactions in the video plausibly adhere to real-world physical laws, contributing to natural-looking dynamics.
\end{enumerate}

\paragraph{Automated Metrics}
We complement our AI and human evaluations with automated metrics that measure video quality and script alignment.
\begin{itemize}[leftmargin=12pt]
    \item \textbf{Standard Quality Metrics.} We adopt several established metrics to assess general video quality, including the \textbf{CLIP Score}~\citep{radford2021learning} for global semantic alignment and a subset metrics (e.g., subject and background consistency, motion smoothness) of \textbf{VBench}~\citep{huang2024vbench}.

    \item \textbf{Visual-Script Alignment (VSA).} A key limitation of standard metrics is that they often measure \emph{whether} content appears but not \emph{when it appears}. To address this, we propose \textbf{Visual-Script Alignment (VSA)}, a novel metric designed to evaluate \textit{temporal-semantic fidelity}. VSA measures whether visual events occur within their designated time intervals as specified by the script. Given a script with $K$ shot units, each with an instruction $I_k$ and time interval $T_k$, and a video with frames $v_t$, VSA is defined as:
    \begin{equation}
        \text{VSA} = \frac{1}{\sum_{k=1}^K |T_k|} \sum_{k=1}^K \sum_{t \in T_k} \text{Sim}\left( \mathcal{E}_{\text{vis}}(v_t), \mathcal{E}_{\text{txt}}(I_k) \right),
    \end{equation}
    where $\mathcal{E}_{\text{vis}}$ and $\mathcal{E}_{\text{txt}}$ are CLIP encoders. A higher score indicates stronger alignment between the video's temporal structure and the script's plan.
\end{itemize}

\section{Experiment}

\subsection{Results of Script Generation}
We compare against representative story visualization and screenplay generation methods:
\begin{itemize}[leftmargin=12pt]
\item {\em MoPS}~\citep{ma2024mops} proposes a modular framework for automated story premise synthesis by decomposing premises into theme, background, persona, and plot modules, then recombining them via a nested dictionary and LLM-based integration to generate diverse, high-quality story foundations.
\item {\em CHAE}~\citep{wang2022chae} enables fine-grained controllable story generation by allowing users to specify characters, their actions, and emotions through a structured input format, enhanced with a copy mechanism and character-wise emotion loss for precise narrative control.
\item {\em SEED-Story}~\citep{yang2025seed} extends multimodal large language models to generate long, coherent narratives interleaved with images, utilizing a multimodal attention sink mechanism to maintain consistency and enable generation beyond training sequence lengths.
\end{itemize}

\begin{table}[t]
\centering
\caption{Script generation evaluation on the \scriptdata~ test set.}
\label{tab:script_eval}
\small
\begin{tabular}{lccccccc}
\toprule
\multirow{3}{*}{\textbf{Method}} & \multicolumn{4}{c}{\textbf{AI Rating (0-5)}} & \multicolumn{3}{c}{\textbf{Human Rating (0-5)}} \\
\cmidrule(lr){2-5} \cmidrule(lr){6-8}
& \em Format & \em Shot & \em Content & \em Narrative & \em Character & \em Dramatic & \em Visual \\
& \em Comp. & \em Division & \em Comp. & \em Coher. & \em Consist. & \em Tension & \em Imagery \\
\midrule
CHAE \small\citep{wang2022chae} & 3.3 & 3.2 & 3.4 & 3.5 & 3.1 & 3.3 & 3.4 \\
MoPS \small\citep{ma2024mops} & 3.2 & 3.1 & 3.3 & 3.4 & 3.0 & 3.2 & 3.3 \\
SEED-Story \small\citep{yang2025seed} & 3.6 & 3.5 & 3.7 & 3.8 & 3.6 & 3.7 & 3.8 \\
\midrule
\textbf{ScriptAgent (SFT only)} & {3.9} & {3.6} & {3.8} & {3.9} & {3.7} & {3.6} & {3.8} \\
\rowcolor{gray!20}
\textbf{ScriptAgent (Full)} & \textbf{4.0} & \textbf{3.9} & \textbf{4.1} & \textbf{4.2} & \textbf{4.0} & \textbf{4.1} & \textbf{4.3} \\
\bottomrule
\end{tabular}
\end{table}

\paragraph{Our proposed ScriptAgent significantly outperforms all baseline methods in generating high-quality cinematic scripts.} As demonstrated in Table~\ref{tab:script_eval}, the full \scriptagent~ model achieves state-of-the-art results across all AI and human evaluation metrics. It shows marked superiority over the strongest baseline, \texttt{MovieAgent}, in structural quality, with improvements of +0.4 points in \textit{Format Compliance}, \textit{Content Completeness}, and \textit{Narrative Coherence}. More importantly, human evaluators confirm its superior creative capabilities, awarding it substantially higher scores in \textit{Visual Imagery} (4.3 vs. 3.8) and \textit{Dramatic Tension} (4.1 vs. 3.7). These results confirm that \scriptagent, trained on our \scriptdata benchmark, effectively bridges the ``semantic gap'' by transforming coarse dialogue into detailed, cinematically expressive, and director-level instructions.

\paragraph{The preference-alignment RL stage is crucial for elevating creative and artistic quality beyond structural correctness.} To isolate the impact of our two-stage training paradigm, we compare the full \scriptagent~ (SFT+RL) with an SFT-only variant. The results show that the SFT-only model already learns the basic script structure, achieving strong scores in \textit{Format Compliance} (3.9) and \textit{Narrative Coherence} (3.9). However, the subsequent GRPO-based preference alignment stage provides significant boosts to more subjective, artistic dimensions. For instance, the score for \textit{Dramatic Tension} improves from 3.6 to 4.1, and \textit{Visual Imagery} rises from 3.8 to 4.3. This finding validates our hypothesis that while SFT is sufficient for structural competence, the RL stage with its hybrid reward function is essential for aligning the model with expert directorial aesthetics, refining its ability to handle nuanced creative elements like pacing and shot composition.

\begin{figure}[t!]
    \centering
    \includegraphics[width=\linewidth]{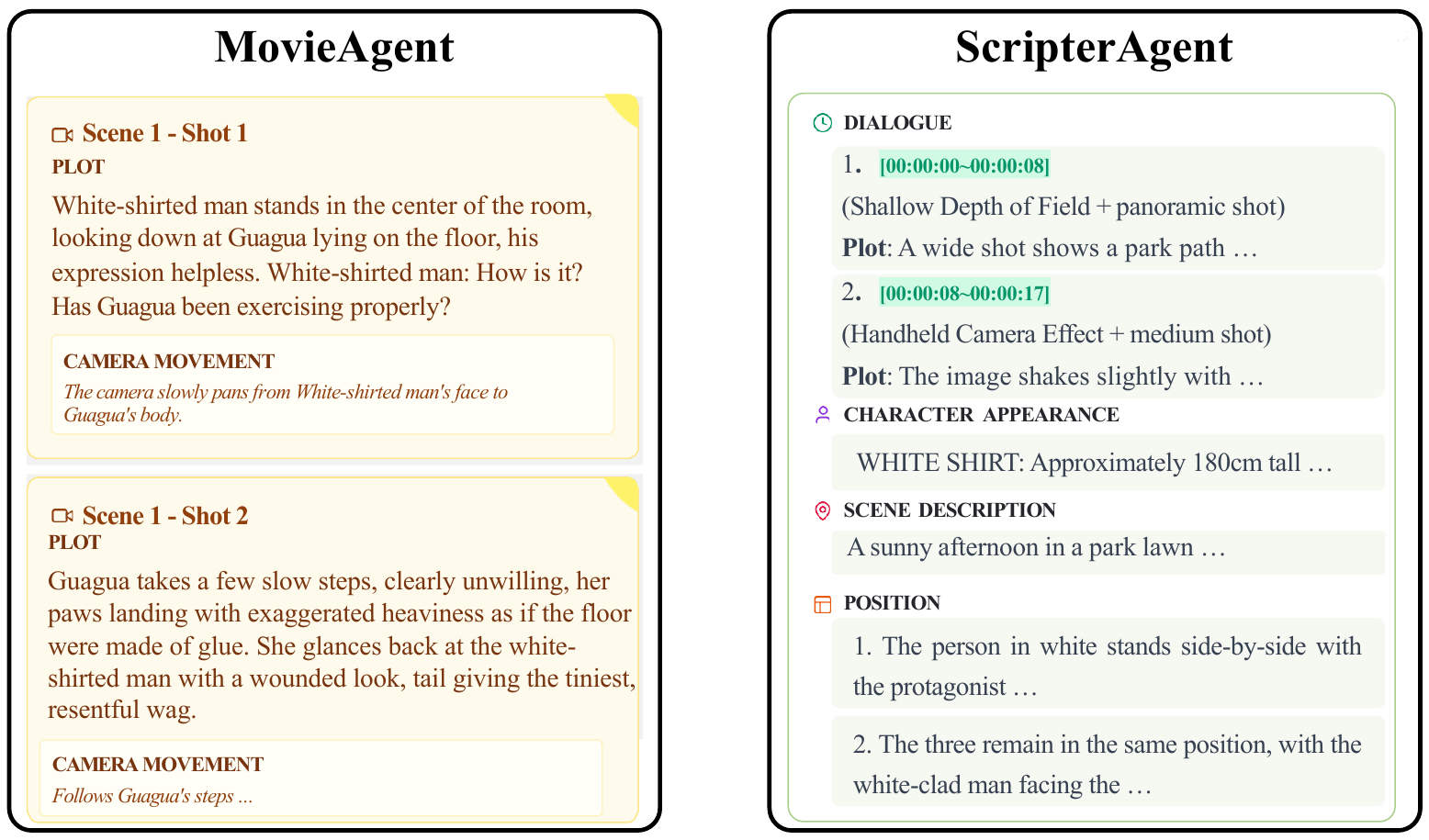}
    \caption{Qualitative comparison: \scriptagent~ vs. MovieAgent~\citep{wu2025automated}.}
    \label{fig:script_comparison}
\end{figure}

\paragraph{Qualitative analysis further highlights \scriptagent's ability to generate a detailed, executable filmmaking plan.} A direct comparison of generated outputs reveals the practical superiority of our approach, as shown in Figure~\ref{fig:script_comparison}. The script from \texttt{MovieAgent} provides a simple plot summary, such as ``Camera slowly pans...''. In stark contrast, the \scriptagent~ output constitutes a complete, fine-grained cinematic blueprint. It specifies precise technical details like camera settings (``Shallow Depth of Field + panoramic shot''), timestamps for synchronization (``[00:00:00~00:00:08]''), detailed character appearance (``Approximately 180cm tall''), atmospheric scene descriptions, and exact character positioning or ``blocking''. This richness confirms that \scriptagent~ successfully generates the professional-quality, executable script needed to guide automated video production, directly addressing the core challenge outlined in our introduction.

\subsection{Results of Video Generation}

We evaluate SOTA text-to-video models supporting voice generation and long-text inputs ($>1000$ tokens):
\begin{itemize}[leftmargin=12pt]
    \item \texttt{Vidu2}~\citep{bao2024vidu} (Shengshu Technology, Tsinghua University): A U-ViT-based model excelling in temporal consistency and generation speed.
    \item \texttt{Seedance1.5-Pro}~\citep{chen2025seedance} (ByteDance Seed): A high-fidelity diffusion model specialized in generating professional-grade videos with enhanced dynamic coherence and visual detail.
    \item \texttt{Kling2.6} (Kuaishou Technology): A Transformer-based generative model capable of synthesizing high-resolution videos with improved motion fluidity and scene understanding.
    \item \texttt{Wan2.6}~\citep{wan2025wan} (Alibaba): A diffusion-based framework synthesizing realistic scenes with intricate details and smooth temporal dynamics.
    \item  \texttt{HYVideo1.5}~\citep{wu2025hyvideo1.5} (Tencent): A large-scale video generation framework featuring a dual-stream diffusion transformer architecture that achieves state-of-the-art performance in prompt following and 4K-resolution synthesis. 
    \item \texttt{Sora2-Pro}~\citep{brooks2024video} (OpenAI): A diffusion transformer model generating high-fidelity, physically plausible videos with complex scenes.
    \item \texttt{Veo3.1} (Google DeepMind): A generative model creating high-resolution (e.g., 1080p) videos with strong cinematic quality and motion coherence.
\end{itemize}

\subsubsection{Results of AI and Human Rating}

\begin{table*}[t]
    \centering
    \caption{Video generation evaluation on the \scriptdata~ test set.}
    \label{tab:video_eval}
    \setlength{\tabcolsep}{0.5pt}
    \begin{tabular}{c ccccc ccccc cc}
    \toprule
    \multirow{3}{*}{\textbf{Model}} & \multicolumn{5}{c}{\textbf{AI Rating (0-5)}} & \multicolumn{5}{c}{\textbf{Human Rating (0-5)}} & \multicolumn{2}{c}{\textbf{Overall Mean}} \\
    \cmidrule(lr){2-6} \cmidrule(lr){7-11} \cmidrule(lr){12-13}
    & \em \small Cam. & \em \small Body & \em \small Visual & \em \small Emotion & \em \small Pace & \em \small Visual & \em \small Script & \em \small Character & \em \small Physical &  \em \small Narrative & \em Avg. & \em Avg. \\
    & \em \small Artic. & \em \small Block. & \em \small Fidelity & \em \small Arc & \em \small Timing & \em \small Appeal & \em \small Faith & \em \small Consist.  & \em \small Law   &   \em \small Coher. & \em AI & \em Human \\
    \midrule
    \multicolumn{13}{c}{\bf Raw Dialogue (w/o \scriptagent)}\\
    \texttt{Vidu2}      & 4.1 & 4.1 & 4.5 & 4.4 & 4.4 & 3.7 & 4.1 & 3.0 & 3.3 & 3.1 & 4.3 & 3.4 \\
    \texttt{Seedance1.5-Pro}     & 4.0 & 4.0 & 4.5 & 4.2 & 4.3 & 3.5 & 3.7 & 3.2 & 3.1 & 3.5 & 4.2 & 3.4 \\
    \texttt{Kling2.6}     & 4.1 & 4.1 & 4.6 & 4.4 & 4.4 & 3.6 & 3.5 & 3.3 & 3.4 & 3.7 & 4.3 & 3.5 \\
    \texttt{Wan2.6}     & 4.2 & 4.2 & 4.7 & 4.4 & 4.4 & 3.5 & 3.2 & 3.1 & 3.7 & 3.4 & 4.4 & 3.4 \\
    \texttt{HYVideo1.5} & 4.0 & 4.0 & 4.5 & 4.3 & 4.3 & 4.0 & 4.2 & 4.1 & 3.8 & 4.1 & 4.2 & 4.0 \\
    \texttt{Sora2-Pro}  & 4.1 & 4.0 & 4.6 & 4.3 & 4.3 & 4.2 & 3.6 & 3.7 & 4.1 & 3.9 & 4.3 & 3.9 \\
    \texttt{Veo3.1}     & 4.0 & 3.9 & 4.4 & 4.4 & 4.3 & 3.9 & 4.0 & 4.1 & 3.9 & 4.0 & 4.2 & 4.0 \\
    \hdashline
    \rowcolor{gray!20}
    Average Score       & 4.1 & 4.0 & 4.5 & 4.3 & 4.3 & 3.8 & 3.8 & 3.5 & 3.6 & 3.7 & 4.2 & 3.7 \\
    \midrule
    \multicolumn{13}{c}{\bf w/ \scriptagent}\\
    \texttt{Vidu2}      & 4.2 & 4.4 & 4.7 & 4.5 & 4.5 & 3.9 & 4.3 & 3.7 & 3.9 & 3.8 & 4.5 & 3.9 \\
    \texttt{Seedance1.5-Pro}     & 4.5 & 4.6 & 4.7 & 4.6 & 4.7 & 4.0 & 4.1 & 4.1 & 3.9 & 4.1 & 4.6 & 4.0 \\
    \texttt{Kling2.6}     & 4.3 & 4.5 & 4.6 & 4.5 & 4.6 & 3.9 & 4.1 & 4.0 & 4.2 & 4.1 & 4.5 & 4.1 \\
    \texttt{Wan2.6}     & 4.4 & \bf4.6 & 4.7 & 4.6 & 4.7 & 4.1 & 4.0 & 3.8 & 4.0 & 3.9 & 4.6 & 4.0 \\
    \texttt{HYVideo1.5} & \bf4.4 & 4.5 & \bf4.8 & 4.5 & \bf4.7 & 4.5 & \bf4.6 & \bf4.4 & 4.2 & \bf4.3 & 4.6 & 4.4 \\
    \texttt{Sora2-Pro}  & 4.1 & 4.4 & 4.7 & 4.5 & 4.6 & \bf4.8 & 4.2 & 4.3 & \bf4.5 & 4.1 & 4.5 & 4.4 \\
    \texttt{Veo3.1}     & 4.1 & 4.4 & 4.5 & \bf4.6 & 4.4 & 4.6 & 4.4 & 4.3 & 4.4 & 4.2 & 4.4 & 4.4 \\
    \hdashline
    \rowcolor{gray!20}
    Average Score       & 4.3 & 4.5 & 4.7 & 4.5 & 4.6 & 4.3 & 4.2 & 4.1 & 4.2 & 4.1 & 4.5 & 4.2 \\
    \bottomrule
    \end{tabular}

\end{table*}

\paragraph{\scriptagent{} enables faithful, temporally coherent video generation by translating dialogue into executable cinematic structure.} Conditioning video models on structured scripts produced by \scriptagent{} yields consistent improvements across all evaluated dimensions (Table~\ref{tab:video_eval}). Aggregated results reveal a uniform uplift: the mean AI rating rises from 4.2 to 4.5, and the mean human rating increases from 3.7 to 4.2 . This efficacy is most pronounced in \textit{Script Faithfulness}, with \texttt{Wan2.6} improving from 3.2 to 4.0 and \texttt{Sora2-Pro} from 3.6 to 4.2, while \texttt{HYVideo1.5} achieves the highest overall fidelity (4.6). Beyond semantic alignment, explicit shot-level blocking instructions enhance fine-grained execution, boosting AI-rated \textit{Pace Timing} and \textit{Body Blocking} by synchronizing motion with scene rhythm. Furthermore, gains in \textit{Character Consistency} and \textit{Narrative Coherence} validate \directoragent's Cross-Scene strategy; by coupling boundary-aware segmentation with frame-anchoring, the framework mitigates identity drift and extends coherent generation beyond single-model limits. Collectively, these results confirm that the domain-informed plans from \scriptagent{} provide essential guidance absent in raw dialogue.

\paragraph{Our evaluation reveals a fundamental trade-off in SOTA models between visual spectacle and strict script adherence.} The results expose a clear divergence in model capabilities that corroborates our third contribution. \texttt{Sora2-Pro} excels in visual impact, securing top scores in \textit{Visual Appeal} (4.8) and \textit{Physical Law} adherence (4.5), making it ideal for high-spectacle generation where realism is paramount. Conversely, \texttt{HYVideo1.5} prioritizes narrative integrity, leading in \textit{Script Faithfulness} (4.6), \textit{Character Consistency} (4.4), and \textit{Narrative Coherence} (4.3). This dichotomy suggests that current video models optimize along different axes: some prioritize perceptual realism, while others better maintain the semantic logic of a storyline when guided by structured scripts. This insight provides valuable guidance for practitioners selecting models for specific filmmaking applications.

\begin{figure*}[t!]
    \centering
    \subfloat[Example1]{
    \includegraphics[width=1.0\linewidth]{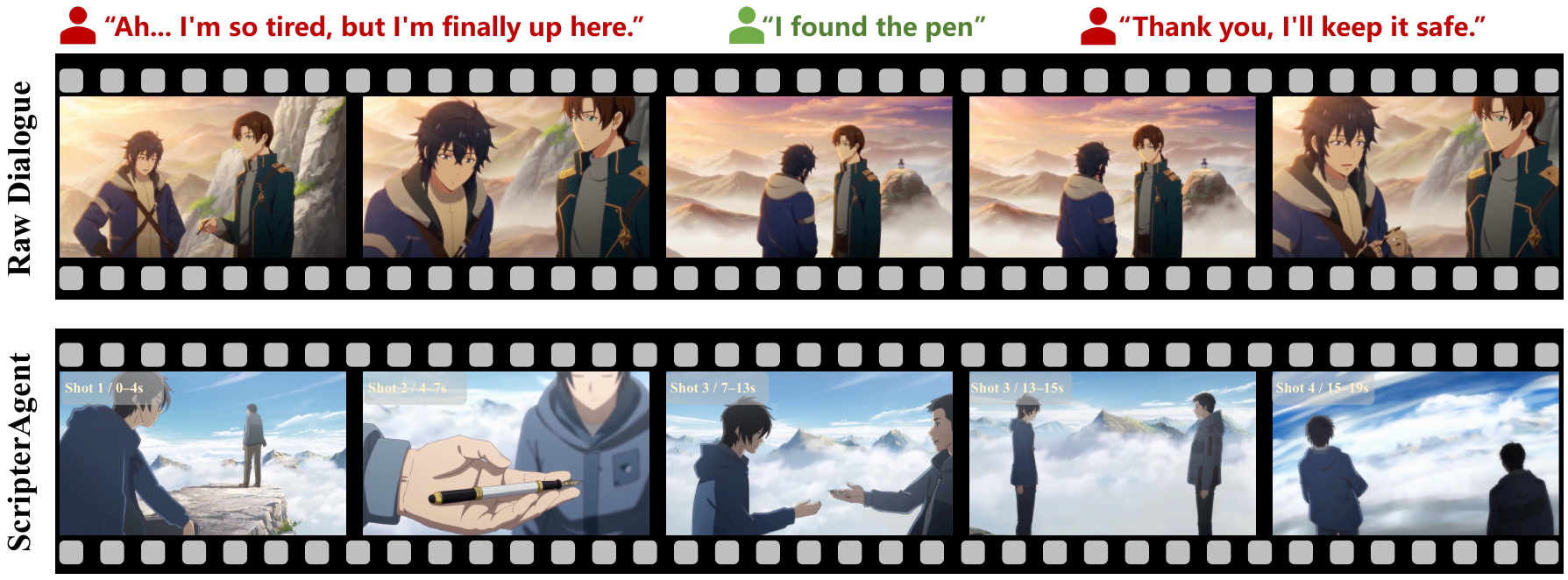}}\\
    \subfloat[Example2]{
    \includegraphics[width=1.0\linewidth]{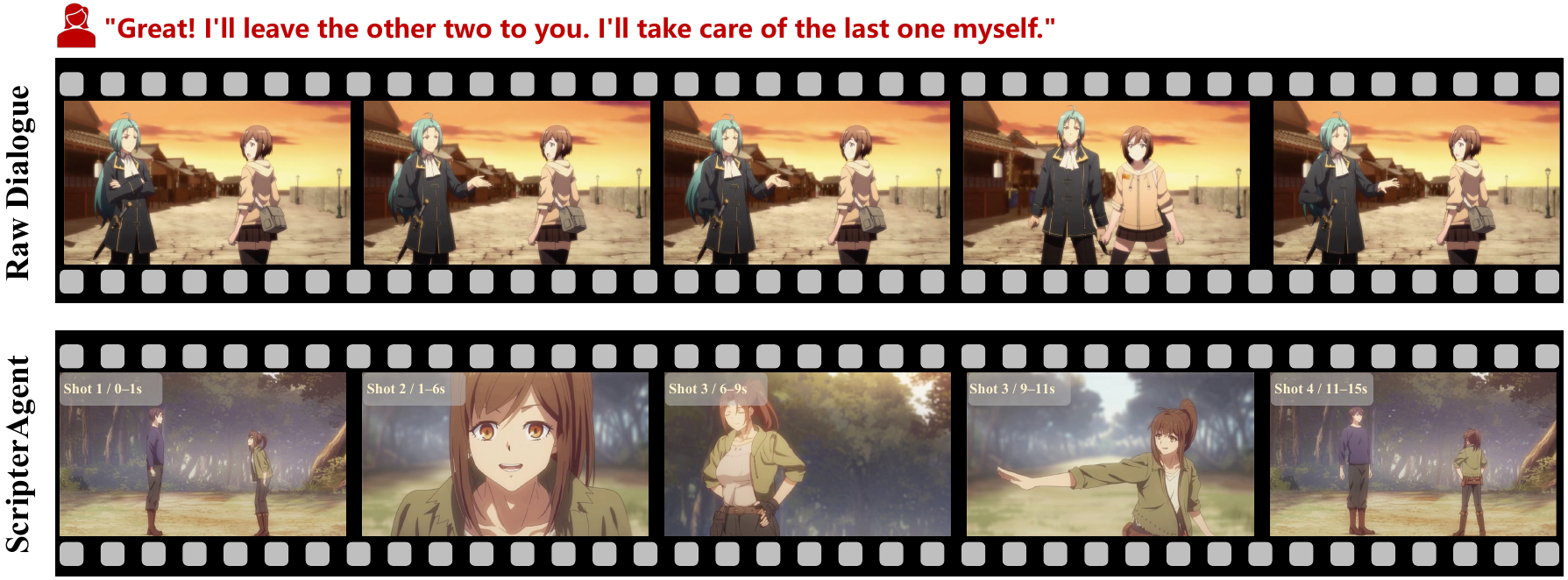}}
    \caption{Examples of video generation using \texttt{Sora2-Pro}.}
    \label{fig:video_cases}
\end{figure*}

\subsubsection{Results of Automatic Metrics}

\begin{table}[t]
    \centering
    \setlength{\tabcolsep}{4pt}
    \caption{Automatic metrics of video generation on the \scriptdata~ test set.}
    \label{tab:video_auto}
    \small
    \begin{tabular}{c cc ccccc}
    \toprule
    \multirow{3}{*}{\textbf{Model}} & \multicolumn{2}{c}{\textbf{Alignment Fidelity}} & \multicolumn{5}{c}{\textbf{VBench Metrics\%} $\uparrow$} \\ 
    \cmidrule(lr){2-3} \cmidrule(lr){4-8}
       & \em CLIP $\uparrow$ & \em VSA $\uparrow$ & \em Subj. Cons. & \em Bg Cons. & \em Mot. Smth. & \em Dyn. Deg. & \em Aesthetic \\ 
    \midrule
    \multicolumn{8}{c}{\bf w/o \scriptagent}\\
    \texttt{Vidu2}          & 42.2 & 48.2 & 92.8 & 93.1 & 94.5 & 46.5 & 48.1 \\
    \texttt{Seedance1.5-Pro}& 43.8 & 50.4 & 93.5 & 93.8 & 95.0 & 48.5 & 49.0 \\
    \texttt{Kling2.6}       & 44.7 & 51.3 & 92.5 & 92.8 & 95.5 & 52.0 & 53.5 \\
    \texttt{Wan2.5}         & 45.3 & 52.1 & 91.5 & 91.8 & 94.9 & 50.2 & 51.0 \\
    \texttt{HYVideo1.5}     & 43.4 & 52.7 & 95.8 & 96.0 & 96.1 & 68.8 & 57.5 \\
    \texttt{Sora2-Pro}      & 44.1 & 48.6 & 90.6 & 91.0 & 96.8 & 75.2 & 59.5 \\
    \texttt{Veo3.1}         & 43.6 & 51.4 & 94.2 & 94.8 & 95.5 & 63.1 & 55.2 \\
    \midrule
    \multicolumn{8}{c}{\bf w/ \scriptagent}\\
    \texttt{Vidu2}          & 43.9 & 50.0 & 94.1 & 94.3 & 95.8 & 49.2 & 50.5 \\
    \texttt{Seedance1.5-Pro}& 45.5 & 52.6 & 94.8 & 95.0 & 96.0 & 51.5 & 51.5 \\
    \texttt{Kling2.6}       & 46.2 & 53.5 & 93.8 & 94.0 & 96.8 & 56.5 & 56.0 \\
    \texttt{Wan2.5}         & 47.2 & 54.1 & 92.8 & 93.1 & 96.2 & 53.8 & 53.4 \\
    \texttt{HYVideo1.5}     & 45.1 & \bf54.8 & \bf97.2 & \bf97.5 & 97.3 & 72.5 & 60.2 \\
    \texttt{Sora2-Pro}      & 46.0 & 50.6 & 91.8 & 92.5 & \bf98.2 & \bf79.5 & \bf62.8 \\
    \texttt{Veo3.1}         & 45.3 & 53.8 & 95.8 & 96.1 & 96.9 & 66.4 & 57.5 \\
    \bottomrule
    \end{tabular}
\end{table}

\paragraph{\textbf{Our VSA metric confirms \scriptagent~ enhances temporal-semantic fidelity.}}
As shown in Table~\ref{tab:video_auto}, conditioning on \scriptagent{} outputs yields consistent gains in our proposed Visual-Script Alignment (VSA) metric. For instance, \texttt{Veo3.1}’s VSA score increases from 51.4 to 53.8, and \texttt{Sora2-Pro}’s from 48.6 to 50.6. \texttt{HYVideo1.5} achieves the highest performance with a VSA score of 54.8. This quantitatively validates that the detailed instructions from \scriptagent{} enable models to better adhere to script semantics and timing. This enhanced alignment is supported by consistent improvements in CLIP scores across all models (Average +1.7), although we note that strict adherence occasionally impacts pure aesthetic scores for some models, suggesting a nuanced trade-off between fidelity and perceptual beauty.

\paragraph{\textbf{Automated metrics reinforce the trade-off between visual dynamism and script adherence.}}
The quantitative data in Table~\ref{tab:video_auto} corroborates our qualitative findings. \texttt{Sora2-Pro} leads in visual spectacle, achieving the highest scores in \textit{Aesthetic} quality (62.8), \textit{Dynamic Degree} (79.5), and \textit{Motion Smoothness} (98.2). In contrast, \texttt{HYVideo1.5} excels at instruction following, posting the highest VSA score (54.8) for temporal alignment, while also leading in \textit{Subject Consistency} (97.2) and \textit{Background Consistency} (97.5), confirming its strength in maintaining narrative integrity over long horizons.

\paragraph{\textbf{\scriptagent~ elicits more dynamic and consistent video generation.}}
Using \scriptagent{} prompts an increase in the \textit{Dynamic Degree} metric across all models. This is highlighted by \texttt{Sora2-Pro}’s increase from 75.2 to 79.5 and \texttt{Kling2.6}'s rise from 52.0 to 56.5. This suggests that explicit action descriptions in the scripts guide models to create more visually complex scenes, moving beyond the static “talking head” outputs often produced from raw dialogue. Furthermore, widespread improvements in \textit{Subject} and \textit{Background Consistency} validate that our cross-scene generation strategy effectively mitigates identity drift.

\section{Related Work}

\paragraph{Video Generation.}
Recent video generation relies on diffusion models~\citep{blattmann2023align,he2022latent,ho2022imagen,khachatryan2023text2video,singer2022make,bao2024vidu,brooks2024video,wan2025wan,wu2025hyvideo1.5} and language models~\citep{hong2022cogvideo,zhan2025lexsembridge,gong2026trace,chang2022maskgit,chang2023muse,kondratyuk2023videopoet,villegas2022phenaki}. 
However, existing video generation systems still face significant limitations in managing long-form narrative coherence, especially when dealing with complex film scripts~\citep{chen2025skyreels}.
Our work addresses these challenges with a cross-scene generation strategy that mitigates fixed-duration constraints.

\paragraph{LLMs for Film Production.}
LLMs are increasingly used to automate film production tasks like scene generation, character planning, and cinematography~\citep{lin2023videodirectorgpt,wei2022emergent}. Systems such as Anim-Director~\citep{li2024anim} help generate storylines and refine scenes, while MovieAgent~\citep{wu2025automated} uses multi-agent collaboration for automated film creation. A key limitation of these models is their reliance on manual input for narrative and cinematographic planning. In contrast, our approach introduces a comprehensive end-to-end pipeline~\citep{huang2025filmaster} that automates scene structuring and planning, reducing manual intervention and ensuring adherence to professional filmmaking standards.

\paragraph{Story Visualization.}
Story visualization maps scripts to visual sequences. Early methods (StoryGAN~\citep{li2019storygan}) produced static, temporally incoherent images. Recent diffusion models (StoryDiffusion~\citep{zhou2024storydiffusion}, Magic-Me~\citep{ma2024magic}) improve temporal consistency and motion but still lack automated high-level planning for cinematography, scene structure, and character interactions, thus requiring manual guidance. We introduce a multi-agent, chain-of-thought (CoT) framework that uses hierarchical reasoning to automate long-form movie generation, ensuring temporal consistency, narrative integrity, and visual appeal over extended durations.

\section{Conclusion}

In this work, we addressed the critical challenge of generating long-form, coherent cinematic videos from sparse dialogue. We proposed a novel, script-centric agentic framework that successfully bridges the semantic gap between high-level narrative concepts and low-level video synthesis. Our system, featuring \scriptagent, \directoragent, and \criticagent, automates the pipeline from dialogue to final video by first generating a detailed, executable cinematic script. The efficacy of our approach is rooted in the high-quality \scriptdata~benchmark and the innovative two-stage (SFT+RL) training of \scriptagent.

Our experiments empirically validate that this script-driven approach universally enhances the narrative coherence, temporal fidelity, and script faithfulness of state-of-the-art video generation models. Our proposed Visual-Script Alignment (VSA) metric provides a quantitative measure of this improvement. Moreover, this study offers the first systematic analysis of the trade-off between visual spectacle and narrative adherence in leading video models, a key insight for the field. This work moves beyond simple prompt-based generation, laying the groundwork for sophisticated automated storytelling systems capable of reasoning about cinematography, pacing, and long-term consistency. Future research could focus on enhancing fine-grained control, such as precise lip synchronization, and developing adaptive models that can dynamically generate content in diverse cinematic styles.

\clearpage

\bibliography{ref}
\bibliographystyle{colm2024_conference}

\clearpage

\appendix

\section{Supplementary Experiment}

\subsection{Additional experiments for Table~\ref{tab:video_eval}}
\begin{table}[htbp]

\setlength{\tabcolsep}{6pt}
\centering
\caption{Video generation evaluation on the \scriptdata~ test set by Qwen3-VL.}
\label{tab:video_eval_qwen}
\begin{tabular}{c ccccc c}
\toprule
\multirow{3}{*}{\textbf{Model}} &
\multicolumn{5}{c}{\textbf{AI Rating (0-5)}} & \multirow{3}{*}{\textbf{Avg.}}\\
\cmidrule(lr){2-6}
& \em\small Cam. & \em\small Body & \em\small Visual & \em\small Emotion & \em\small Pace & \\
& \em\small Artic. & \em\small Block. & \em\small Fidelity & \em\small Arc & \em\small Timing & \\
\midrule
\multicolumn{7}{c}{\bf Raw Dialogue (w/o \scriptagent)}\\
\texttt{Vidu2}        & 3.9 & 3.8 & 4.3 & 4.3 & 4.1 & 4.1\\
\texttt{Seedance1.5-Pro}       & 4.0 & 4.1 & 4.4 & 4.4 & 4.2 & 4.2\\
\texttt{Kling2.6}       & 3.8 & 3.8 & 4.2 & 4.2 & 4.0 & 4.0\\
\texttt{Wan2.6}       & 4.1 & 4.0 & 4.5 & 4.5 & 4.3 & 4.3\\
\texttt{HYVideo1.5}   & 3.8 & 3.7 & 4.2 & 4.2 & 4.0 & 4.0\\
\texttt{Sora2-Pro}    & 3.8 & 3.8 & 4.3 & 4.3 & 4.1 & 4.1\\
\texttt{Veo3.1}       & 3.8 & 3.8 & 4.2 & 4.3 & 4.0 & 4.0\\
\hdashline
Average Score         & 3.9 & 3.9 & 4.3 & 4.3 & 4.1 & 4.1\\
\midrule
\multicolumn{7}{c}{\bf w/ \scriptagent}\\
\texttt{Vidu2}        & 4.6 & 4.6 & 4.4 & 4.7 & 4.7 & 4.6\\
\texttt{Seedance1.5-Pro}       & 4.6 & 4.5 & 4.4 & 4.7 & 4.7 & 4.6\\
\texttt{Kling2.6}       & 4.6 & 4.6 & 4.4 & 4.7 & 4.8 & 4.6\\
\texttt{Wan2.6}       & 4.6 & 4.6 & 4.4 & 4.7 & 4.7 & 4.6\\
\texttt{HYVideo1.5}   & 4.6 & 4.6 & 4.4 & 4.7 & 4.7 & 4.6\\
\texttt{Sora2-Pro}    & 4.6 & 4.5 & 4.3 & 4.6 & 4.7 & 4.5\\
\texttt{Veo3.1}       & 4.6 & 4.5 & 4.4 & 4.6 & 4.7 & 4.6\\
\hdashline
Average Score         & 4.6 & 4.6 & 4.4 & 4.7 & 4.7 & 4.6\\
\bottomrule
\end{tabular}

\end{table}

\subsection{Ablation Study on Scripting and Agentic Components}
\label{sec:appendix_ablation}

\subsubsection{Experimental Configuration}
We evaluate the incremental impact of our framework across four stages:
\begin{itemize}
    \item \textbf{Baseline (Raw Dialogue):} Generating videos directly from dialogue prompts without ScripterAgent.
    \item \textbf{w/ Script Only:} Conditioning on ScripterAgent's detailed scripts, but generating as a single long-horizon clip.
    \item \textbf{w/ Script + Segment:} Implementing shot-aware segmentation but without the frame-anchoring consistency mechanism.
    \item \textbf{Full Agent (Ours):} The complete pipeline featuring script-conditioning, segmentation, and \textit{Frame-Anchoring} for visual continuity.
\end{itemize}

\begin{table*}[t]
    \resizebox{\textwidth}{!}{%
    \setlength{\tabcolsep}{3.3pt}
    \begin{tabular}{l ccccc ccccc ccccc}
    \toprule
    \multirow{3}{*}{\textbf{Setting}} & \multicolumn{5}{c}{\textbf{Gemini 2.5 Pro}} & \multicolumn{5}{c}{\textbf{Qwen 3 VL}} & \multicolumn{5}{c}{\textbf{GLM 4.6 V}} \\
    \cmidrule(lr){2-6} \cmidrule(lr){7-11} \cmidrule(lr){12-16}
    & \em \small Cam. & \em \small Body & \em \small Visual & \em \small Emotion & \em \small Pace & \em \small Cam. & \em \small Body & \em \small Visual & \em \small Emotion & \em \small Pace & \em \small Cam. & \em \small Body & \em \small Visual & \em \small Emotion & \em \small Pace \\
    & \em \small Artic. & \em \small Block. & \em \small Fidelity & \em \small Arc & \em \small Timing & \em \small Artic. & \em \small Block. & \em \small Fidelity & \em \small Arc & \em \small Timing & \em \small Artic. & \em \small Block. & \em \small Fidelity & \em \small Arc & \em \small Timing \\
    \midrule
    \multicolumn{16}{c}{\textbf{Wan2.6}} \\
    Baseline & 4.2 & 4.2 & 4.7 & 4.4 & 4.4 & 4.1 & 4.0 & 4.5 & 4.5 & 4.3 & 4.0 & 4.1 & 4.2 & 4.0 & 4.1 \\
    w/ Script Only & 4.2 & 4.3 & 4.7 & 4.5 & 4.5 & 4.3 & 4.2 & 4.5 & 4.6 & 4.4 & 4.2 & 4.4 & 4.3 & 4.2 & 4.3 \\
    w/ Script+Seg. & 4.3 & 4.4 & 4.7 & 4.5 & 4.6 & 4.5 & 4.4 & 4.4 & 4.6 & 4.6 & 4.4 & 4.6 & 4.4 & 4.4 & 4.4 \\
    \textbf{Full Agent} & \textbf{4.4} & \textbf{4.6} & \textbf{4.7} & \textbf{4.6} & \textbf{4.7} & \textbf{4.6} & \textbf{4.6} & \textbf{4.4} & \textbf{4.7} & \textbf{4.7} & \textbf{4.5} & \textbf{4.8} & \textbf{4.5} & \textbf{4.5} & \textbf{4.5} \\
    \midrule
    \multicolumn{16}{c}{\textbf{HYVideo1.5}} \\
    Baseline & 4.0 & 4.0 & 4.5 & 4.3 & 4.3 & 3.8 & 3.7 & 4.2 & 4.2 & 4.0 & 4.1 & 3.8 & 4.2 & 4.1 & 4.0 \\
    w/ Script Only & 4.1 & 4.2 & 4.6 & 4.4 & 4.4 & 4.1 & 4.0 & 4.3 & 4.4 & 4.2 & 4.3 & 4.0 & 4.5 & 4.3 & 4.3 \\
    w/ Script+Seg. & 4.3 & 4.3 & 4.7 & 4.4 & 4.6 & 4.4 & 4.3 & 4.4 & 4.6 & 4.5 & 4.5 & 4.1 & 4.7 & 4.5 & 4.5 \\
    \textbf{Full Agent} & \textbf{4.4} & \textbf{4.5} & \textbf{4.8} & \textbf{4.5} & \textbf{4.7} & \textbf{4.6} & \textbf{4.6} & \textbf{4.4} & \textbf{4.7} & \textbf{4.7} & \textbf{4.6} & \textbf{4.2} & \textbf{4.9} & \textbf{4.6} & \textbf{4.6} \\
    \bottomrule
    \end{tabular}%
    }
    \centering
    \caption{Ablation results on Wan2.6 and HYVideo1.5 across three AI evaluators. All scores are on a 0-5 scale.}
    \label{tab:detailed_ablation}
\end{table*}

\subsubsection{Quantitative Results Analysis} We evaluate the generated videos across five cinematic dimensions using a panel of three advanced LLMs as critics: Gemini 2.5 Pro, Qwen 3 VL, and GLM 4.6 V. As shown in Table~\ref{tab:detailed_ablation}, the \textbf{Full Agent} consistently outperforms baselines across all metrics and backbone models (Wan2.6 and HYVideo1.5).

The ablation study reveals a clear functional decoupling of our agents. {\em Visual Descriptive Fidelity} and {\em Kinetic Body Language \& Blocking} show marked improvements with the introduction of \textit{w/ Script Only}, attributing fine-grained visual control to the \scriptagent{}. Conversely, the \directoragent{} is shown to be critical for temporal dimensions; the implementation of shot-aware segmentation and frame-anchoring drives scores in {\em Narrative Pacing \& Timing} and {\em Cinematic Camera Articulation} to their highest levels. The high agreement among the three diverse critic models strongly corroborates the validity of these improvements.

\section{Details on Evaluation Metrics}
\label{appendix:metrics}

In this section, we detail the human evaluation criteria used in our experiments and all scores are assigned on a $0$--$5$ scale.

\subsection{Human Evaluation for Script Generation}

Annotators read the source dialogue and the generated script, then assign $0$--$5$ scores for each metric according to the following criteria.

\textbf{Character Portrayal Consistency.}  
This metric evaluates whether each character’s personality, speaking style, and behavior remain coherent and believable throughout the script.
\begin{itemize}[leftmargin=12pt]
    \item \textit{Score 0 (Invalid or Off-Topic):} The script is largely unrelated to the given dialogue, missing key characters or scenes, or is so fragmented that character portrayal cannot be meaningfully judged.
    \item \textit{Score 1 (Completely Inconsistent):} Characters frequently change personality, tone, or goals without any narrative justification, making them feel arbitrary or incomprehensible.
    \item \textit{Score 2 (Severe Inconsistencies):} Major contradictions in speech style or behavior appear, and characters often feel unstable or unconvincing.
    \item \textit{Score 3 (Noticeable Variations):} Core traits are roughly maintained, but several noticeable shifts in dialogue or actions break immersion.
    \item \textit{Score 4 (Good Consistency):} Characters are generally consistent in personality and voice, with only minor deviations that do not seriously affect believability.
    \item \textit{Score 5 (Perfect Consistency):} Each character maintains a stable and well-defined identity, with coherent speech and behavior in all scenes.
\end{itemize}

\textbf{Dramatic Tension \& Rhythm.}  
This metric measures how well the script builds, sustains, and releases dramatic tension, and whether the pacing of events feels natural and engaging.
\begin{itemize}[leftmargin=12pt]
    \item \textit{Score 0 (Invalid or No Story):} The text does not form a recognizable story (e.g., mostly noise, repetition, or unrelated fragments), so dramatic structure cannot be evaluated.
    \item \textit{Score 1 (No Dramatic Structure):} The script lacks recognizable conflicts or turning points, and the pacing feels random or monotonous.
    \item \textit{Score 2 (Weak and Uneven Tension):} Some conflicts exist but are poorly set up or resolved, with abrupt or dragging sections that hurt engagement.
    \item \textit{Score 3 (Basic but Uneven Rhythm):} The overall arc is understandable, but there are noticeable pacing issues (e.g., rushed climaxes or overlong trivial scenes).
    \item \textit{Score 4 (Good Dramatic Arc):} Conflicts, climaxes, and resolutions are well-structured, with generally appropriate pacing and emotional build-up.
    \item \textit{Score 5 (Strong and Compelling Tension):} The script offers a clear and engaging dramatic curve, with well-timed beats that sustain viewer interest throughout.
\end{itemize}

\textbf{Visual Imagery \& Cinematic Expressiveness.}  
This metric assesses how vividly the script conveys visual information and how well it uses cinematic language (shots, staging, atmosphere) to support filming.
\begin{itemize}[leftmargin=12pt]
    \item \textit{Score 0 (Not Filmable / Unreadable):} The script is so incomplete, disorganized, or off-topic that it provides no meaningful basis for imagining shots or scenes.
    \item \textit{Score 1 (Vague and Non-Visual):} Descriptions are extremely abstract, offering almost no cues for camera work, staging, or visual composition.
    \item \textit{Score 2 (Sparse Visual Guidance):} Only a few scattered visual hints are provided; most scenes are difficult to imagine concretely on screen.
    \item \textit{Score 3 (Basic Visual Clarity):} Key scenes and actions are described clearly enough to imagine, but shot types or cinematic details remain generic.
    \item \textit{Score 4 (Good Cinematic Guidance):} The script includes clear descriptions of scenes, actions, and rough shot intentions, enabling straightforward visualization.
    \item \textit{Score 5 (Highly Cinematic and Filmable):} The script shows rich visual imagination and appropriate use of film language, making it easy to translate into professional storyboards.
\end{itemize}

\subsection{Human Evaluation for Video Generation}

Annotators watch each generated video with access to the corresponding script (or dialogue) as reference, and then give $0$--$5$ scores according to the following guidelines.

\textbf{Visual Appeal.}  
This metric evaluates the realism, aesthetic quality, and rendering stability of the video.
\begin{itemize}[leftmargin=12pt]
    \item \textit{Score 0 (Missing or Corrupted Video):} The video cannot be properly viewed (e.g., file error, almost entirely static or black frames), so visual quality is not assessable.
    \item \textit{Score 1 (Severe Artifacts):} Heavy distortions, glitches, or unrecognizable content dominate most frames, making the video difficult to watch.
    \item \textit{Score 2 (Poor Visual Quality):} Frequent flickering, unstable textures, and obvious inconsistencies, though core content is still somewhat interpretable.
    \item \textit{Score 3 (Moderate Issues):} Overall content is clear, but noticeable artifacts in textures, lighting, or motion transitions reduce visual quality.
    \item \textit{Score 4 (Good Visual Quality):} Generally realistic and stable visuals with only minor distortions or occasional flickering that do not strongly affect viewing.
    \item \textit{Score 5 (High-Quality Rendering):} Smooth, realistic, and aesthetically pleasing visuals, comparable to professionally produced footage, with minimal visible artifacts.
\end{itemize}

\textbf{Script Faithfulness.}  
This metric assesses how accurately the video follows the intended script or ScriptAgent-generated plan in terms of scenes, actions, and plot progression.
\begin{itemize}[leftmargin=12pt]
    \item \textit{Score 0 (No Relation or No Reference):} The video has almost no identifiable connection to the given script (or the script is missing/invalid), making faithfulness impossible to judge.
    \item \textit{Score 1 (Almost Unrelated):} The video bears little or no resemblance to the script, missing key scenes, characters, or events.
    \item \textit{Score 2 (Major Deviations):} Some script elements appear, but important settings, actions, or turning points are missing, misplaced, or heavily altered.
    \item \textit{Score 3 (Partial Alignment):} The main storyline can be recognized, but there are noticeable inaccuracies or omissions in details and shot arrangement.
    \item \textit{Score 4 (Good Faithfulness):} Most key moments and settings match the script, with only minor deviations that do not significantly affect overall story understanding.
    \item \textit{Score 5 (Almost Perfect Adherence):} The video closely follows the script in both content and structure, accurately reflecting specified scenes, actions, and plot beats.
\end{itemize}

\textbf{Narrative Coherence.}  
This metric measures whether the video forms a logically consistent and easy-to-follow story, with reasonable scene transitions and pacing.
\begin{itemize}[leftmargin=12pt]
    \item \textit{Score 0 (No Discernible Narrative):} The video is so fragmented, repetitive, or random that no coherent storyline or temporal order can be inferred.
    \item \textit{Score 1 (Completely Incoherent):} Scenes appear random and disconnected, with no understandable storyline or causal relations.
    \item \textit{Score 2 (Frequent Confusion):} Some story intention is visible, but abrupt cuts and illogical transitions make the plot hard to follow.
    \item \textit{Score 3 (Partially Coherent):} A basic story can be inferred, yet inconsistencies or awkward pacing often disrupt narrative flow.
    \item \textit{Score 4 (Well-Structured Story):} The video presents a mostly clear and coherent narrative, with only minor issues in pacing or transitions.
    \item \textit{Score 5 (Fully Coherent and Engaging):} The plot develops smoothly and logically, with seamless scene transitions and a clear, engaging storytelling arc.
\end{itemize}

\textbf{Character Consistency.}  
This metric evaluates whether characters maintain a stable identity and appearance across the entire video.
\begin{itemize}[leftmargin=12pt]
    \item \textit{Score 0 (Characters Not Identifiable):} Human or main characters are almost entirely missing, severely deformed, or indistinguishable, so consistency cannot be meaningfully evaluated.
    \item \textit{Score 1 (Completely Inconsistent):} Character faces, bodies, or outfits change drastically between shots, making them hard to recognize as the same person.
    \item \textit{Score 2 (Severe Inconsistencies):} Frequent noticeable changes in facial features, clothing, or proportions, even if some continuity is preserved.
    \item \textit{Score 3 (Moderate Variations):} Characters are generally recognizable, but variations in appearance or style appear multiple times and affect immersion.
    \item \textit{Score 4 (Good Consistency):} Character identity and look are mostly stable, with only small visual fluctuations that do not seriously disrupt continuity.
    \item \textit{Score 5 (Perfect Consistency):} Characters keep a highly stable appearance and identity across all shots, making them visually coherent throughout the video.
\end{itemize}

\textbf{Physical Law Adherence.}  
This metric assesses whether motions and interactions in the video roughly follow real-world physical laws, contributing to natural-looking dynamics.
\begin{itemize}[leftmargin=12pt]
    \item \textit{Score 0 (Motion Not Assessable):} The video is nearly static, heavily corrupted, or too abstract, so physical plausibility of motion cannot be reasonably judged.
    \item \textit{Score 1 (Highly Unrealistic):} Objects or characters frequently move in impossible ways (e.g., floating, severe limb distortions) without any narrative justification.
    \item \textit{Score 2 (Many Violations):} Multiple obvious physical inconsistencies (e.g., unnatural collisions, gravity-defying motion), although not in every scene.
    \item \textit{Score 3 (Partially Plausible):} Most movements are acceptable, but there are several noticeable errors in weight, balance, or continuity of motion.
    \item \textit{Score 4 (Mostly Realistic):} Characters and objects move in a generally natural way, with only minor physical anomalies that are easy to overlook.
    \item \textit{Score 5 (Highly Plausible Physics):} Motion and interactions appear smooth and physically convincing, with no obvious violations of basic physical laws.
\end{itemize}

\subsection{AI-Based Rating Metrics}

To complement objective metrics, we use LLM-based evaluators to score both \emph{script generation} and \emph{video generation} on a $0$–$5$ scale aligned with film production standards. For scripts, Gemini-2.5-Pro focuses on structural and logical correctness; for videos, Gemini-2.5-Pro focuses on perceptual and technical audio-visual quality. Below we show the full prompts and the detailed scoring rubric for each dimension.

\begin{promptbox}[Script Generation Evaluation Prompt]{nblue}
You are a professional film director and script supervisor. Your task is to evaluate the quality of a generated shooting script based on the provided coarse-grained dialogue and context.

\textbf{Input Data:}
\begin{itemize}[leftmargin=12pt]
    \item \textbf{Source Dialogue:} \{Insert Origin Dialogue Here\}
    \item \textbf{Generated Script:} \{Insert Generated JSON Script Here\}
\end{itemize}

\textbf{Evaluation Criteria:}
Please score the generated script on a scale of 0 to 5 for each of the following dimensions.  
For each dimension, use the following general guideline:

\begin{itemize}[leftmargin=12pt]
    \item \textit{Score 0:} Completely unusable or fails the requirement.
    \item \textit{Score 1:} Very poor quality; severe issues in most parts.
    \item \textit{Score 2:} Clearly below acceptable quality; many issues.
    \item \textit{Score 3:} Acceptable but with noticeable issues.
    \item \textit{Score 4:} Good quality with only minor issues.
    \item \textit{Score 5:} Excellent quality; no meaningful issues.
\end{itemize}

Then, judge each dimension more concretely as follows:

\begin{enumerate}
    \item \textbf{Format Compliance (0–5):}  
    Does the output strictly follow the required JSON format? Are all key fields (Shot Type, Camera Movement, Description, etc.) present and correctly structured?
    \begin{itemize}[leftmargin=12pt]
        \item \textit{0:} Not valid JSON or completely ignores the requested schema.
        \item \textit{1:} Severe structural errors; many missing or malformed fields.
        \item \textit{2:} Multiple structural problems; only partially follows the schema.
        \item \textit{3:} Mostly follows the schema but with some missing fields or minor format issues.
        \item \textit{4:} Fully follows the schema with only very small formatting inconsistencies.
        \item \textit{5:} Perfectly formatted JSON with all fields correctly present and structured.
    \end{itemize}

    \item \textbf{Shot Division Rationality (0–5):}  
    Is the script segmented into shots reasonably? Do the shot breaks align with narrative beats and emotional shifts without being too fragmented or too long?
    \begin{itemize}[leftmargin=12pt]
        \item \textit{0:} No meaningful shot division; essentially a single block or random splitting.
        \item \textit{1:} Very unreasonable segmentation; shots break the flow and ignore story structure.
        \item \textit{2:} Many inappropriate shot boundaries; frequent over- or under-segmentation.
        \item \textit{3:} Basic correspondence to narrative beats, but with several awkward or suboptimal shot splits.
        \item \textit{4:} Mostly well-aligned with emotional and narrative shifts, with only minor segmentation issues.
        \item \textit{5:} Shot division is highly reasonable, closely following narrative and emotional structure throughout.
    \end{itemize}
    \item \textbf{Content Completeness (0–5):}  
    Does the script provide rich, actionable details for filming? Does it supplement necessary visual information that was missing in the source dialogue?
    \begin{itemize}[leftmargin=12pt]
        \item \textit{0:} Almost no additional visual or staging information beyond the raw dialogue.
        \item \textit{1:} Very sparse detail; crucial information for filming is largely missing.
        \item \textit{2:} Some useful details, but important aspects (scene, actions, camera) remain underspecified.
        \item \textit{3:} Contains enough information to stage the scene, but important visual details are still missing.
        \item \textit{4:} Generally rich and specific, covering most necessary visual, spatial, and action details.
        \item \textit{5:} Highly complete and specific, providing clear and thorough guidance for key visual choices.
    \end{itemize}
    \item[4.] \textbf{Narrative Coherence (0–5):}  
    Is the sequence of shots logically connected? Does the visual storytelling flow smoothly and match the context of the dialogue?
    \begin{itemize}[leftmargin=12pt]
        \item \textit{0:} Completely incoherent sequence; shots appear random and unrelated to the dialogue.
        \item \textit{1:} Very confusing progression; frequent contradictions or abrupt jumps.
        \item \textit{2:} A rough story is visible, but there are many logical gaps, contradictions, or unnatural transitions.
        \item \textit{3:} Overall story is understandable, but several transitions or details break the narrative flow.
        \item \textit{4:} Mostly coherent and smoothly flowing narrative with only minor inconsistencies.
        \item \textit{5:} Fully coherent, well-structured visual narrative that aligns closely with the dialogue context.
    \end{itemize}
\end{enumerate}
\textbf{Output Format:}  
Return the result in the following JSON format:
\begin{verbatim}
{
  "Format Compliance": [Score],
  "Shot Division Rationality": [Score],
  "Content Completeness": [Score],
  "Narrative Coherence": [Score]
}
\end{verbatim}
\end{promptbox}

\begin{promptbox}[Video Generation Evaluation Prompt]{norange}

You are an expert AI Film Critic and Cinematographer with deep expertise in visual storytelling, camera techniques, and cinematic language. Your task is to evaluate the video's cinematic quality and adherence to complex directorial instructions.

\textbf{Input Data:}
\begin{itemize}[leftmargin=12pt]
    \item \textbf{Reference Script:} \{Insert Reference Script Here\}
    \item \textbf{Generated Video:} \{Video File to be Evaluated\}
\end{itemize}

\textbf{Evaluation Criteria:}
Please score the video on a scale of 0.0 to 5.0 for each of the following cinematic dimensions. You can assign ANY decimal score (e.g., 2.3, 3.7, 4.2). The integer benchmarks (0, 1, 2, 3, 4, 5) serve as REFERENCE POINTS for quality boundaries.

\textbf{IMPORTANT:} Simple dialogue videos with minimal movement should receive LOW scores (1-2). HIGH scores (4-5) are reserved ONLY for videos demonstrating sophisticated cinematic techniques.

\begin{itemize}[leftmargin=12pt]
    \item \textit{Score 0:} Completely fails the requirement; no evidence of the evaluated quality.
    \item \textit{Score 1:} Minimal/default quality; severe problems throughout.
    \item \textit{Score 2:} Basic quality; many noticeable issues.
    \item \textit{Score 3:} Competent/functional quality; acceptable but uninspired.
    \item \textit{Score 4:} Advanced/dynamic quality; well-executed with minor issues.
    \item \textit{Score 5:} Master-level quality; exceptional execution indistinguishable from professional cinema.
\end{itemize}

Then, judge each dimension more concretely as follows:

\begin{enumerate}
    \item \textbf{Cinematic Camera Articulation (0.0–5.0):}  
    Evaluates the sophistication and intentionality of camera work, including movement, framing transitions, and visual storytelling techniques.
    \begin{itemize}[leftmargin=12pt]
        \item \textit{0:} Completely static camera; single unchanging framing; feels like a frozen screenshot.
        \item \textit{1:} Predominantly static with occasional accidental shifts; simple linear zoom with no artistic purpose; AI default setting.
        \item \textit{2:} Simple panning/tilting with uniform speed; basic zoom uncorrelated with narrative; abrupt transitions.
        \item \textit{3:} Clear shot variety (Wide/Medium/Close-up); camera movements motivated by action; demonstrates basic cinematic grammar but mechanical execution.
        \item \textit{4:} Purposeful dynamic techniques (handheld shake, tracking shots, focal shifts); smooth transitions aligned with narrative peaks; camera positioning creates visual tension.
        \item \textit{5:} Exceptional sophisticated camera language (fluid handheld, crane shots, dolly moves); perfect composition; focal shifts precisely timed; every camera decision serves narrative purpose.
    \end{itemize}

    \item \textbf{Kinetic Body Language \& Blocking (0.0–5.0):}  
    Assesses physical performance quality, spatial relationships (blocking), and how bodies express narrative and emotion.
    \begin{itemize}[leftmargin=12pt]
        \item \textit{0:} Characters completely static like mannequins; no gestures or facial movement; zero physicality.
        \item \textit{1:} Only basic lip movement; stiff/repetitive gestures; no spatial repositioning; AI-generated feel with no human quality.
        \item \textit{2:} Simple gestures lacking fluidity; mechanical walking; spatial relationships accidental; gestures don't match emotional context.
        \item \textit{3:} Characters move with basic purpose (A to B); contextually appropriate but unspecific gestures; basic blocking present; believable but not expressive.
        \item \textit{4:} Highly specific actions (running and stopping at precise point, leaning forward, catching breath); intentional spatial blocking creates tension; body language evolves through scene.
        \item \textit{5:} Every action precise and motivated; micro-movements (fidgeting, weight shifts); complex sequences; blocking choreographed to perfection reflecting power dynamics; culturally-specific gestures.
    \end{itemize}

    \item \textbf{Visual Descriptive Fidelity (0.0–5.0):}  
    Measures how accurately visual output matches script descriptions, including character appearance, clothing textures, environmental details, and lighting.
    \begin{itemize}[leftmargin=12pt]
        \item \textit{0:} Characters look random or change appearance; environment blank/incoherent; lighting broken.
        \item \textit{1:} Characters vaguely human but bear no resemblance to descriptions; generic clothing contradicts script; lighting ignores time-of-day cues.

    \end{itemize}
    \begin{itemize}[leftmargin=12pt]
        \item \textit{2:} Gender/age match but specific features wrong; clothing category correct but textures/colors incorrect; environment thematically correct but lacks details.
        \item \textit{3:} Major descriptors match (gender, age, clothing style); environment includes key elements but simplified; lighting matches time-of-day but lacks detailed effects.
        \item \textit{4:} Characters closely match detailed descriptions (hair style, clothing textures, body type); environment shows specific details (pavement texture, metal railings); sophisticated lighting with proper shadows.
        \item \textit{5:} Photorealistic precision with every descriptor present; micro-details (fabric wrinkles, button placement); environmental lighting interacts realistically; atmospheric depth; indistinguishable from high-end cinematography.
    \end{itemize}
    
    \item \textbf{Emotional Arc \& Micro-Expressions (0.0–5.0):}  
    Evaluates range and authenticity of emotional performance, including facial expressions, emotional transitions, and psychological subtext.
    \begin{itemize}[leftmargin=12pt]
        \item \textit{0:} Faces blank/frozen/mask-like; no visible emotional state; characters appear lifeless.
        \item \textit{1:} Single unchanging expression; clearly looped frames; no emotional reaction to events; contradicts narrative context; robotic feel.
        \item \textit{2:} One or two basic emotions expressed simplistically/exaggerated; abrupt transitions; lacks nuance; doesn't align with dialogue tone; cartoonish.
        \item \textit{3:} Emotional states generally match dialogue; at least one clear shift; basic facial movements (eyebrow raises, mouth changes); recognizable but generic; lacks micro-expressions.
        \item \textit{4:} Multiple distinct emotional states with clear transitions (laugh→serious→questioning); nuanced details (eyebrow furrows, eye contact changes); micro-expressions present; emotional intensity varies appropriately; feels ``acted'' not generated.
        \item \textit{5:} Rich layered emotional journey; complex arcs (playful→realization→concern→resolve); exceptional micro-expressions (1-2 frame fleeting expressions); emotions blend naturally; character-consistent and psychologically motivated; subtext visible; indistinguishable from professional actor performance.
    \end{itemize}

    \item \textbf{Narrative Pacing \& Timing (0.0–5.0):}  
    Assesses whether video executes clear narrative structure with appropriate timing, action sequencing, and rhythmic flow matching script's story beats.
    \begin{itemize}[leftmargin=12pt]
        \item \textit{0:} Video incoherent; no discernible beginning/middle/end; actions occur randomly; timing completely broken.
        \item \textit{1:} Duration wildly mismatches dialogue; actions in wrong order or omitted; no logical flow; feels like random clips stitched together.
        \item \textit{2:} Length approximately matches but internal pacing off; key actions happen at wrong times; some narrative beats present but sequencing confused; rhythm monotonous or chaotic.
        \item \textit{3:} Duration matches dialogue; basic narrative sequence present (setup→event→conclusion); actions in correct order; pacing acceptable but lacks dynamic variation; competent but uninspired.
        \item \textit{4:} Dialogue and action timing precisely synchronized; clear purposeful structure (setup→action→escalation→resolution); pacing creates rhythm; timing builds/releases tension appropriately; action sequences follow believable physics.
        \item \textit{5:} Perfect narrative timing with cinematic rhythm; three-act structure compressed into scene; actions timed with precision to the second; rhythmic variation creates emotional texture; timing builds and releases tension masterfully; pacing feels inevitable and organic; indistinguishable from professionally edited film.
    \end{itemize}
\end{enumerate}

\textbf{Output Format:}
Return ONLY a JSON structure with decimal scores (0.0-5.0), detailed reasoning for each dimension (referencing which benchmarks the video falls between), Final Cinematic Grade (average of all 5 scores), and Overall Assessment.

\textbf{Scoring Reminders:}
\begin{itemize}[leftmargin=12pt]
    \item Use decimal precision (e.g., 2.3, 3.7, 4.5) to distinguish quality levels
    \item Reference integer benchmarks but don't feel limited to them
    \item Explain in reasoning which benchmarks the video falls between and why
    \item Simple dialogue videos with minimal movement should score 1.0-2.5
    \item Only sophisticated, cinema-quality videos should receive scores of 4.0-5.0
\end{itemize}
\end{promptbox}
\twocolumn

\end{document}